\documentclass[conference]{IEEEtran}
\usepackage{cite}
\usepackage{amsmath,amssymb,amsfonts}
\usepackage{algorithmic}
\usepackage{graphicx}
\usepackage{textcomp}
\usepackage{xcolor}

\usepackage{dsfont}
\usepackage[inline]{enumitem}
\usepackage{graphics}
\usepackage{float}
\usepackage{adjustbox}
\usepackage{hyperref}

\def\BibTeX{{\rm B\kern-.05em{\sc i\kern-.025em b}\kern-.08em
    T\kern-.1667em\lower.7ex\hbox{E}\kern-.125emX}}

\DeclareMathOperator{\Ima}{Im}

\DeclareMathOperator{\Out}{Out}
\DeclareMathOperator{\f1}{f1}
\DeclareMathOperator{\ap}{ap}
\DeclareMathOperator{\of1}{of1}
\newcommand{\ree}{\mathds{R}}

\begin{document}

\title{Anomaly detection with partitioning overfitting autoencoder ensembles}

\author{
\IEEEauthorblockN{Boris Lorbeer}
\IEEEauthorblockA{\textit{Technische Universit\"at Berlin} \\
Berlin, Germany \\
lorbeer@tu-berlin.de}
\and
\IEEEauthorblockN{Max Botler}
\IEEEauthorblockA{\textit{Technische Universit\"at Berlin} \\
Berlin, Germany \\
botler@tu-berlin.de}
}

\maketitle

\begin{abstract}
    In this paper, we propose POTATOES (Partitioning OverfiTting AuTOencoder
    EnSemble), a new method for unsupervised outlier detection (UOD). More
    precisely, given any autoencoder for UOD, this technique can be used to
    improve its accuracy while at the same time removing the burden of tuning
    its regularization. The idea is to not regularize at all, but to rather
    randomly partition the data into sufficiently many equally sized parts,
    overfit each part with its own autoencoder, and to use the maximum over
    all autoencoder reconstruction errors as the anomaly score. We apply our
    model to various realistic datasets and show that if the set of inliers is
    dense enough, our method indeed improves the UOD performance of a given
    autoencoder significantly. For reproducibility, the code is made available
    on github so the reader can recreate the results in this paper as well as
    apply the method to other autoencoders and datasets.
\end{abstract}

\begin{IEEEkeywords}
anomaly detection, artificial intelligence, autoencoders, ensembles
\end{IEEEkeywords}

\section{Introduction}
The machine learning task that we are concerned with in this paper is
unsupervised outlier detection (UOD): We are given a dataset in which the vast
majority of members adhere to a certain pattern and only a very small minority
might not (the outliers), but there are no labels available indicating this
minority, not even part of it, and we want to find a way to automatically
single out this minority. Note that in this paper we are only interested in the
situation where
\begin{enumerate*}[label=(\roman*)]
    \item the dataset contains {\em both} the in- {\em and} the outliers and
    \item we are given all the data at once, i.e. we don't have to find a
        method that detects outliers in future data, all data of interest is
        already present.
\end{enumerate*}

Outlier detection (or anomaly detection, this paper uses the words {\em outlier} and
{\em anomaly} synonymously) has become a very popular research topic in machine
learning. One reason is the large variety of methods from numerous subfields of
machine learning that can be applied. There is hardly any discipline that has
not been beneficial to outlier detection.

But arguably most conducive to the rise of the field of anomaly detection is
the broad space of real world applications like predictive maintenance, fraud
detection, quality assurance, network intrusion detection, and data
preprocessing for other machine learning methods such as cleaning data before
training supervised models.

Many current applications of AI like computer vision and natural language
processing have to deal with datasets of high dimension $d$. It is a well known
fact, see e.g.\ \cite{Scholkopf1998}, \cite{Goodfellow},  and \cite{Cayton},
that most of those datasets $D$ have their data points located along a
submanifold $S$ that is of lower, often much lower, dimension $u$ than that of
the ambient space itself, $u\ll d$. In those situations, one popular method of
outlier detection (OD) proceeds as follows: First, one uses manifold learning to
approximate the data with the image of an open subset $U\subset \ree^u$ under a
continuous function: $f: U \to \ree^d$, and sets $S = \Ima f$. Next one
presumes that this $S$ is actually the ``correct'' description of the dataset
$D$ in the sense that a point $p\in D$ would deviate from $S$ only either due
to regular noise (smaller deviation) or due to $p$ actually being an anomaly
(larger deviation). Let us now denote the Euclidean distance between two points
$x,y\in\ree^d$ by $\|x-y\|_2$ and define the distance between a point $p$ and a
submanifold $S$ as $d(p,S)=\inf_{s\in S}\|s-p\|_2$. Then, if those presumptions
above were true, the distances $d(p,S)$ could be considered outlier scores: the
larger the distance, the more likely the point $p$ is an outlier.

One frequently applied technique for obtaining the function $f$ is the
deployment of an autoencoder, see \cite{Chalapathy}. An autoencoder is a
(usually deep) neural network, that takes the points $p\in D$ as input and
tries to reproduce them as output. I.e., it is a function $au:\ree^d\to\ree^d$.
If the goal is to approximate a lower dimensional submanifold, the autoencoder
will have an internal layer with fewer units then the input or output, a
bottleneck. If the submanifold should have dimension $u$, the number of units
in this internal layer will be set to $u$. The activations of the units in the
bottleneck represent a $u$-dimensional vector, which is called the {\em code}
or the {\em latent representation} of the input $p$. The part of the
autoencoder that maps the input to the code, $enc: \ree^d\to\ree^u$, is called
the {\em encoder} and the part that maps the code to the output,
$dec:\ree^u\to\ree^d$, is called the {\em decoder}. Note, that the autoencoder,
is thus the composition of the encoder and the decoder, $au = dec\circ enc$.
After the autoencoder $au$ has been trained, the decoder $dec$ is used as the
above map $f$ defining the submanifold $S$, i.e.\ $S = \Ima dec$.  Furthermore,
the {\em reconstruction error} $\|ae(p)-p\|_2$ is used as approximation for
$d(p,S)$, which has the very convenient effect of freeing us from the
nontrivial task of actually computing $d(p,S)$.

However, it is usually not clear how well $\Ima f$ should approximate the
dataset. If the manifold is chosen to be very flexible such that all points in
the dataset are approximated, outliers will not be distinguishable from
inliers, and we would probably get lots of false negatives. If, on the other
hand, the manifold is chosen to be less flexible, then it might fail to
describe the manifold properly and we would obtain lots of false positives. In
the case of deep neural networks, the flexibility of the approximating
submanifold is controlled by regularization. So a central question is how to
find the right amount of regularization of autoencoders that are used for OD.

To address this problem, we have developed a method to improve the anomaly
detection capabilities of any given autoencoder. We propose a new ensemble
based method that does not require the model designer to spend lots of time
tuning the regularization. The only requirement is that the used autoencoder
model can overfit the dataset. Apart from being easier to construct, it also
gives better results than standard OD autoencoder models, even if those have
been carefully tuned using the knowledge of the true outliers in the dataset,
which is usually not available in practice.

Another important fact to point out is that, even though this paper focuses on
autoencoder based models, the ensemble method described here might also be
applicable to other OD methods that use some form of distance to an
approximating submanifold as the outlier score.

\section{Related Work}
The number of papers dedicated to OD is huge. Here, we will only give a selection
of the publications that we deem most relevant to the idea of the current paper.

Good survey papers are for instance \cite{Hodge}, \cite{Chandola}, or
\cite{Zimek}.  A more recent comparative evaluation is described in
\cite{Domingues}, which, however, does not mention any deep learning based
methods. This is complemented by other papers, e.g.\ \cite{Chalapathy} or
\cite{Kwon}.

The idea of ensemble models is ubiquitous in machine learning and has also
found lots of applications in OD. For example, an ensemble of one-class support
vector machines \cite{Scholkopf} has been proposed in \cite{Perdisci}. It is
applied to network intrusion detection and creates an ensemble by extracting
many different feature spaces from the payload by using 2-grams with different
distances between the two tokens. The perhaps most famous example of model
ensembles in UOD is the isolation forest, see \cite{Liu}.

Using autoencoders and especially ensembles of autoencoders to detect anomalies
is a very popular technique. Lots of papers are dedicated to this idea. E.g.\
the authors of \cite{Chen} use several standard methods of creating ensembles,
like bagging and many autoencoder models with randomly connected units. In
\cite{Mirsky} autoencoder ensemble OD is used for network intrusion detection.
Here, the input features are split into several subsets each of which is fed to
its separate autoencoder. This is the basic idea of other papers, too, e.g.\
\cite{Khan}.

There are also many variants of the supervised approach, i.e.\ considering OD
as a classification task, and thus presuming that labeled data is available.
Examples can be found e.g.\ in \cite{Yaping} or \cite{Qi}.

\section{POTATOES}
\subsection{The regularization trade-off}\label{subsecTradoff}
\begin{figure*}[ht]
    \centering
    \includegraphics[width=\textwidth]{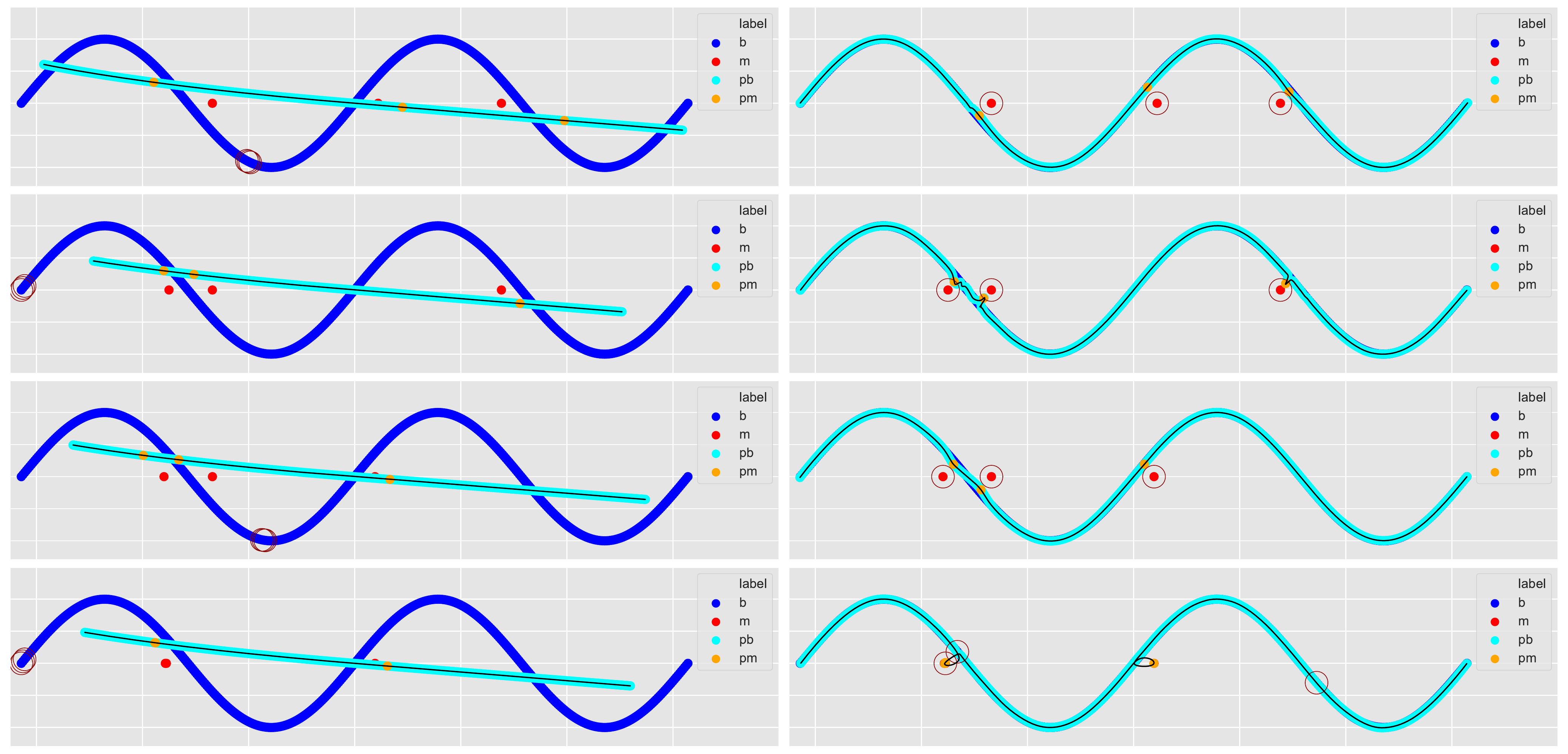}
    \caption{2D Example for the regularization trade-off}
    \label{ae2d}
\end{figure*}
This paper makes use of the well-known method of outlier detection (OD) via
submanifold approximation. Let $D=\{p_i\}_{i=1}^N$ be a set of $N$
$d$-dimensional data points $p_i\in \ree^d$. In many situations, this dataset
is positioned near a submanifold of dimension $u < d$, see
\cite{Scholkopf1998}, \cite{Goodfellow}, and \cite{Cayton}. The task is now to
find this submanifold and to use it as a representative for the data $D$, so
that we can take the distance of any point $p\in D$ to this submanifold as a
measure for the likelihood of this point to be an outlier, i.e.\ as outlier
score. For this to work, the submanifold should be a rather ``smooth''
approximation. One extreme would be a linear sub-plane obtained with PCA, which
is often not flexible enough for properly adapting to the data. Think for
example of data on a sphere. Another extreme would be a strongly nonlinear
submanifold that meets every point in $D$. Here, all $p\in D$ would have an
outlier score of zero, leaving us with no way to differentiate between in- and
outliers. The goal is to find a manifold learning algorithm with a cost
function which favors the optimal ``stiffness'' in the approximating
submanifold that penalizes nonlinearities just the right amount such that the
resulting submanifold will be drawn to where the bulk of the data, the inliers,
is situated, while not bothering to get near isolated data points, the
outliers.

An example for this situation is shown in Figure \ref{ae2d}, where we have used
autoencoder-based UOD as described in the introduction. Each row belongs to a
different dataset, while the columns belong to different autoencoders: in the
first column we see the results of a regularized autoencoder and in the second
column the results of a nonregularized, overfitting autoencoder. The two
dimensional datasets always contain blue inliers on the sine curve (in the
legend denoted by {\bf b} for {\em benign} points), and three red outliers at
different positions in each row (in the legend denoted by {\bf m} for {\em
malign} points). The cyan points ({\bf pb}: {\em predicted benign} points) and
the orange points ({\bf pm}: {\em predicted malign} points) are the output of
the autoencoder for the benign and malign data points, resp. The predicted
points are all positioned on the decoder image of the latent space given by the
black line. The red circles are drawn around the input points that have the
three largest outlier scores. It is obvious that the regularized autoencoder
fails in all four cases to approximate the benign data because it is not
flexible enough. Using the reproduction error as outlier score, taking the
three points with the largest outlier score as our anomalies would result in
three false positives, as indicated by the red circles, and, of course, also in
three false negatives. The overfitting autoencoder, on the other hand, is
approximating the data much better. However, as can be seen particularly
clearly in the fourth row, overfitting can result in an approximation also of
the outliers which then again results in false detections. Intuitively spoken,
regularization controls the ``stiffness'' in the approximating submanifold.
Note, that in the forth row two of the randomly chosen outliers are very close
to each other and look like one single point.

\subsection{Method Details}
In the introduction section we have described a well-known popular technique of
using autoencoders for UOD and in Section \ref{subsecTradoff} we have discussed
common problems with its application. We now propose POTATOES (Partitioning
OverfiTting AuTOencoder EnSemble), a method that both improves the UOD
performance of any given autoencoder, and avoids the above regularization
issues.

To construct POTATOES, we first have to determine how large a cluster of
outlier points can be for those points to still be considered outliers. If we
have a single isolated point, this should clearly be considered an outlier. Two
points being near each other but together isolated from the rest of the
dataset, might still be considered both outliers. But what is the maximum size
$c$ of an outlier cluster? This can only be answered by a domain expert.
Thus, it should be left as a configuration parameter. Once the value of $c$ is
determined, the dataset $D$ of size $N$ is partitioned into $k=c+1$ equally
sized parts. If $k$ divides $N$, each part has size $N/k$, otherwise their
sizes will differ by at most one. Recall the definition of a {\em partition of
a set}: the collection of sets $\{P_1,\ldots,P_k\}$ is a partition of $D$ if:
\begin{align}
    \bigcup_{i=1}^k P_i & = D \\
    P_i \cap P_k        & = \varnothing,\;\mbox{for}\; i\ne k.
\end{align}
Note, that with this partition, for each outlier cluster $C$ there will be at
least one $P_i$ such that $C\cap P_i = \varnothing$.

Next, choose an autoencoder model that is capable of overfitting. Then create
$k$ copies $ae_1, \ldots, ae_k$ of this autoencoder and overfit each of them to
another partition, i.e.\ overfit $ae_i$ to $P_i$. This will result in $k$
submanifolds $S_1,\ldots,S_k$.

\begin{figure}[ht]
    \centering
    \includegraphics[width=0.7\columnwidth]{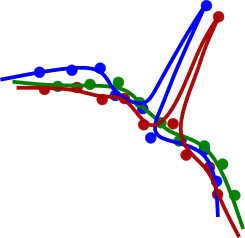}
    \caption{Example of a POTATOES ensemble of three autoencoders fitted to a
    dataset with an outlier cluster of size two}
    \label{potk3}
\end{figure}

An example is provided in Figure \ref{potk3}. Here we choose $c=2$, which means
the displayed dataset has one outlier cluster containing two outliers. The data
is cut into three partitions, depicted by the three colors red, green, and
blue. The fitting of three belonging autoencoders results in three pertinent
submanifolds. 
We see that each inlier has a very short distance to any of the submanifolds.
This is because each inlier has in its close neighborhood far more than three
other datapoints so it is very likely that each submanifold gets close to it.
However, for the outliers the situation is different: since there are less data
points in the outlier cluster than there are partitions, there will be at least
one partition that doesn't contain a member of the outlier cluster and thus not
all submanifolds will get near the outlier cluster and thus each outlier will
have at least one submanifold to which the distance is large.

\begin{figure*}[ht]
    \centering
    \includegraphics[width=\textwidth]{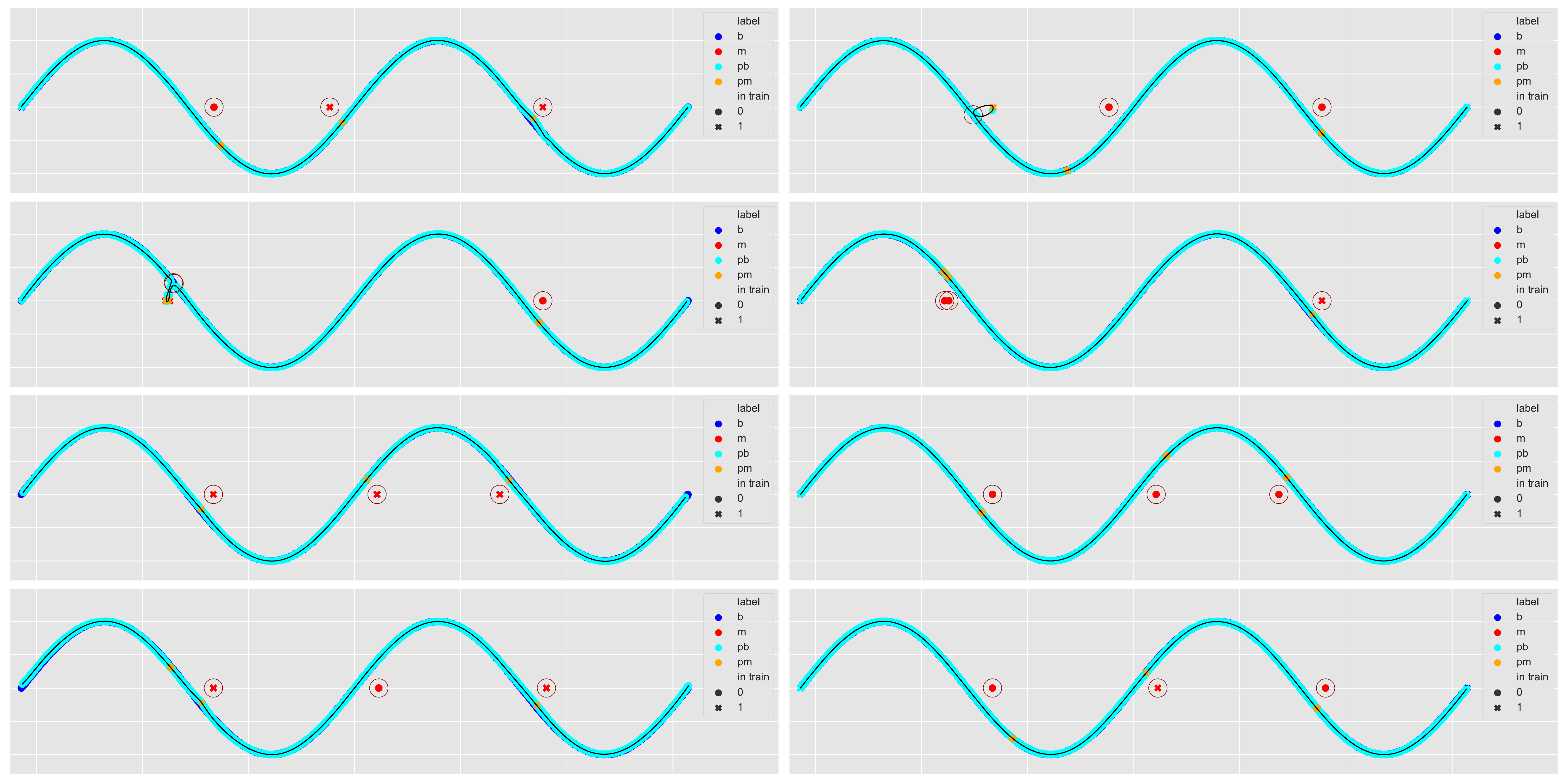}
    \caption{2D Example Application of POTATOES}
    \label{pot2d}
\end{figure*}

This leads to the construction of the outlier score for POTATOES: Let $r_i(p)$
be the reconstruction error of $p\in D$ computed by the autoencoder $ae_i$.
Then for each point $p\in D$, the POTATOES outlier score function $s_{POT}$ is
defined as:
\begin{equation}\label{potOs}
    s_{POT}(p) \triangleq \max_{i=1,\ldots k} r_i(p).
\end{equation}

Note, that with POTATOES it is not necessary to think about any of those
methods that try to fight overfitting just the right amount, such as
regularization parameters, the right choice for early stopping, or constraining
the model complexity. We just have to overfit.

Moreover, As will be shown in the evaluation section, POTATOES significantly
increases the UOD performance of the autoencoder it is based on.

Furthermore, although POTATOES consists of an ensemble of $k$ autoencoders, it
is about as fast as a single autoencoder, since each ensemble member runs only
on one $k$th of the data. Of course, sometimes, overfitting takes more epochs
to converge then regularized models, which then would increase the runtime of
POTATOES. However, the $k$ ensemble autoencoders can be run in parallel,
increasing the speed $k$-fold.

Note, however, that this method only works if there is sufficient data
available such that in each inlier region of the dataset there are data points
of all the partitions $P_i$ located.

While the viewpoint adopted here is that of POTATOES being a method
constructing a set of flexible data manifolds that are ``dense'' near the
inliers and ``sparse'' at the outliers, one could also look at it as another
method of performing UOD via $k$ nearest neighbors ($k$-NNb). But keep in mind
that $k$-NNb based UOD algorithms have at best a complexity of $N\log N$ in the
size $N$ of the dataset, with quadratic complexity being more realistic,
especially for higher dimensional data. POTATOES, on the other hand, has
complexity only linear in $N$.

Lets apply POTATOES to a 2D example, see Figure \ref{pot2d}. As in Figure
\ref{ae2d}, we consider a two-dimensional dataset with the inlier positioned on
the sine curve and three outliers randomly chosen elsewhere. Again, each row
belongs to a different dataset with different outliers. However, now the
columns belong to the members of the POTATOES ensemble with $k=2$. The legend
shows again the same color coding in the label section as in Figure \ref{ae2d},
but now the ``in train'' section has been added. This new section shows that we
use crosses for the points that are contained in the partition of this
autoencoder, i.e.\ are in its training set, and bullets for the points that are
not. Clearly, most of the times, the overfitting autoencoders approximate the
inliers well but don't get to the outliers. However, in two cases, the second
plot in the first row and the first plot in the second row, the autoencoder is
flexible enough to also reach the outliers. Nevertheless, since in each case
there is a second autoencoder that does not contain this outlier in its
training set, the outlier is still recognized as such by POTATOES. This results
in assigning the highest outlier score to the three outliers in each of those
four cases.

\section{Experimental Results}
\begin{figure*}[ht]
    \centering
    \includegraphics[width=\textwidth]{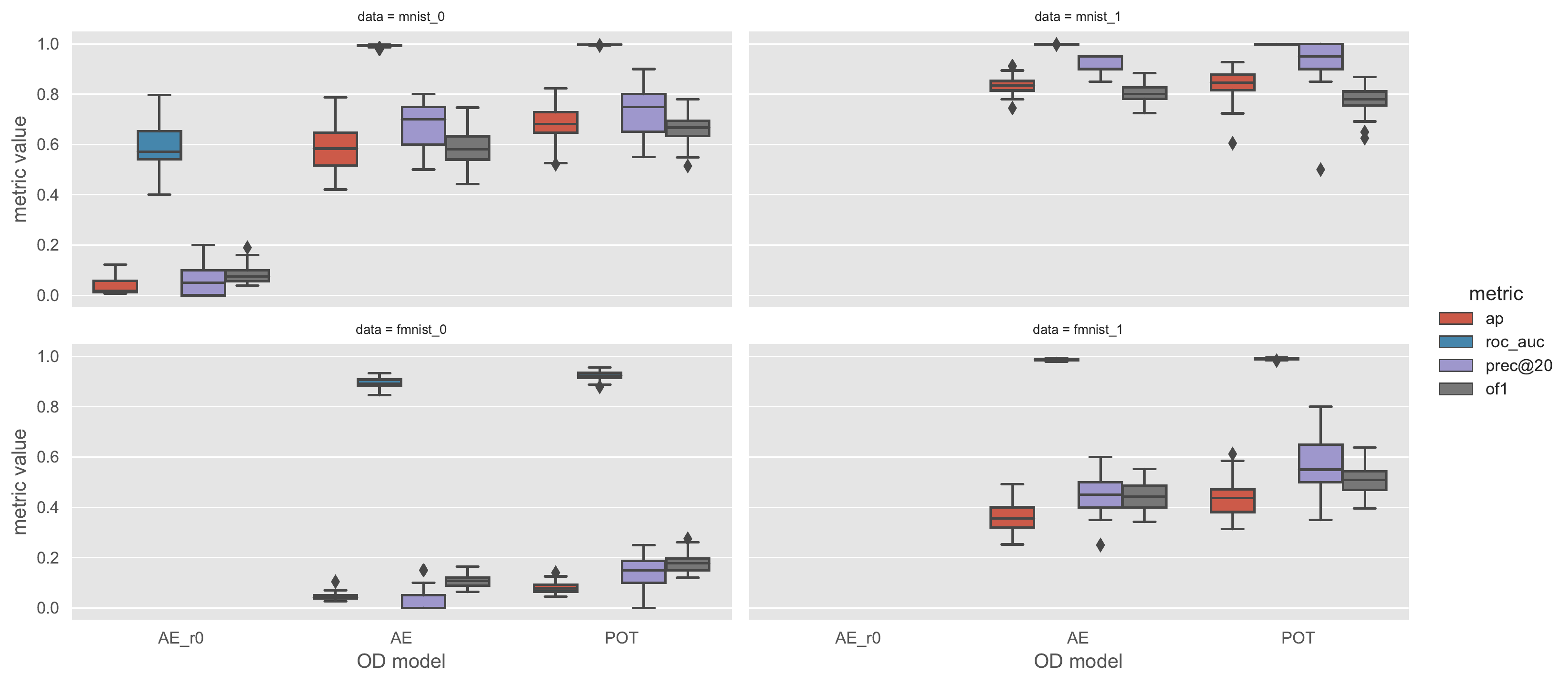}
    \caption{Evaluation of POTATOES}
    \label{evals}
\end{figure*}
For the evaluations in this section, we used the python libraries tensorflow
\cite{tf}, Keras \cite{keras}, and sklearn \cite{sklearn} for model training
and inference, and seaborn \cite{seaborn} for the plots. The source code for
the evaluations in this paper can be found at
https://github.com/snetbl/potatoes. The model is evaluated for the following
datasets. First, we take the well-known MNIST dataset \cite{mnist}, label all
the images containing the zero digit as inliers, and sample from the remaining
nine digits randomly some images and label them as outliers. The amount of
outliers is such that in the combined set $D$ of inliers and outliers, the
portion of outliers is half a percent (i.e.\ $\frac{|\Out(D)|}{|D|} = 0.005$).
We create 50 such datasets, which have all the same inliers but differ by the
randomly sampled outliers. This group will be denoted by mnist\_0. This
procedure is repeated, but this time the inliers are not the images with the
digit zero but the images with the digit one, resulting in two groups mnist\_0
and mnist\_1 of 50 datasets each. Next, the MNIST dataset is replaced by the
Fashion MNIST dataset \cite{fmnist} and the above process is repeated, which
creates two further groups fmnist\_0 and fmnist\_1 (the Fashion MNIST class
labeled 0 comprises the t-shirt/top images, the class labeled 1 the trousers).
In the end, this leaves us with four groups of 50 datasets each.

The metrics used in this evaluation are the {\em ROC area under the curve}
(roc\_auc), the {\em average precision} (ap), {\em optimal F1 score} (of1), and
{\em precision at 20} (prec@20). The of1 metric is computed as follows: if
$s(p)$ denotes the outlier score of the point $p$, and $\f1(D, M)$ is the F1
score of choosing the members of $ M\subset D$ as outliers, then $\of1(D)$ is
given by:
\begin{equation}
    \of1(D) = \max_{p\in D}\;\f1(D, \{q\in D: s(p) < s(q)\}),
\end{equation}
presuming that $M=D$ does not make any sense.

Depending on the amount of inliers, the outlier ratio of 0.005 results in a bit
more than 30 outliers, which means that there are on average between three and
four outliers from each of the nine outlier classes. Presuming that those are
placed near each other, we expect the size of outlier clusters to be more or
less of this size, and thus choose as the POTATOES ensemble size $k=5$. Recall
that choosing $k$ too large might deplete the inlier portions in the $k$ parts
of the partition too much, so we use $k=5$ as a compromise.

Our choice for the original autoencoder, to which we apply the POTATOES
optimization, is a vanilla deep symmetric convolutional architecture with a
latent dimension of 32. The details can be found in the github repository
\cite{githubPot} for this paper. We compare the original autoencoder with the
POTATOES version of it. As described above, while we don't use any
regularization for the POTATOES ensemble, the original autoencoder does get
regularized. This regularization was tuned using the knowledge of the real
labels of the data, giving the regularized autoencoder an advantage that it
would not have in most realistic situations, since labels are usually not
available.

The evaluation results are shown in Figure \ref{evals} and, for the AP metric,
in Table \ref{evalTab}. Here, the regularized autoencoder is abbreviated with
AE and the POTATOES model with POT.

\begin{table}[htbp]
\caption{Mean values of the AP metric}
\begin{center}
\begin{tabular}{|l| r r|}
\hline
              &       AE  &      POT\\
\hline         
    mnist\_0  & 0.579516  & {\bf 0.682226}\\
    mnist\_1  & 0.836213  & {\bf 0.844658}\\
    fmnist\_0 & 0.044366  & {\bf 0.083286}\\
    fmnist\_1 & 0.359578  & {\bf 0.434675}\\
\hline
\end{tabular}
\label{evalTab}
\end{center}
\end{table}

\begin{figure}[ht]
    \centering
    \includegraphics[width=\columnwidth]{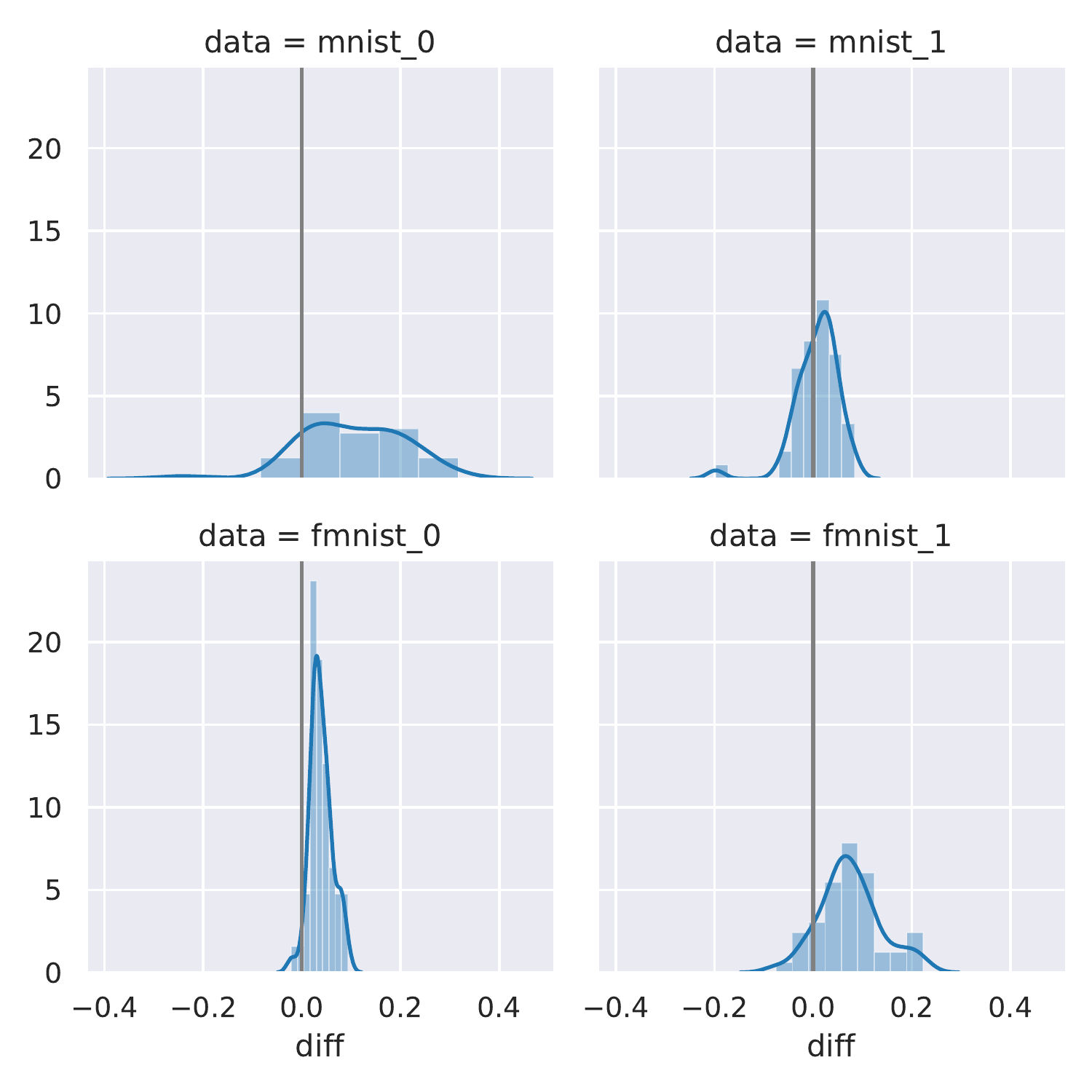}
    \caption{Histograms of the AP differences $\{a_i\}$ between POTATOES and
    its regularized autoencoder}
    \label{diffHists}
\end{figure}

We also ran evaluations for the unregularized version AE\_r0 of AE. The results
for AE\_r0 are only shown for the mnist\_0 dataset, and, because of its poor
performance, has been excluded in further evaluations. Those values clearly
show the potential of POTATOES. To see whether the improvements are actually
significant, we have conducted paired t-tests for the AP metric. For the
application of a paired t-test the following conditions have to be satisfied:
Let $\{D_i\}_{i=1}^t$ be a group of datasets on which we compare the two
models.  Further, let $\ap_{POT}(D), \ap_{AE}(D)$ be the AP values of POTATOES
and the regularized autoencoder on the dataset $D$, resp. Then the differences
$\{a_i\in\ree: a_i = \ap_{POT}(D_i)-\ap_{AE}(D_i)\}_{i=1}^t$ have to be
\begin{enumerate*}[label=(\roman*)]
    \item independent and identically distributed and
    \item samples from a normal distribution.
\end{enumerate*}
The differences $\{a_i\}_{i=1}^t$ are clearly independent in each of our four
dataset groups since the samplings of the outliers in the creation of the
datasets are independent. Next, to check for normality, both the Shapiro-Wilk
and the Kolmogorov-Smirnov test are applied. Figure \ref{diffHists} shows the
histograms of the differences for the four groups of datasets.
Neither Shapiro-Wilk nor Kolmogorov-Smirnov reject the null hypothesis of
normality for any of these four cases, so we can apply the paired two-sample
t-test. The p-values for the four groups are shown in Table \ref{pttp}.
\begin{table}[htbp]
\caption{The p-values of the paired t-tests for each of the four dataset groups}
\begin{center}
\begin{tabular}{|l| r r r r|}
\hline
      &  mnist\_0 &  mnist\_1 & fmnist\_0 & fmnist\_1 \\
\hline                                       
    p & 5.758e-09 & 2.746e-01 & 1.688e-16 & 3.489e-11\\
\hline
\end{tabular}
\label{pttp}
\end{center}
\end{table}
Apart from the mnist\_1 datasets, POTATOES always performs significantly better
than its autoencoder of the same architecture.

\begin{figure}[ht]
    \centering
    \includegraphics[width=\columnwidth]{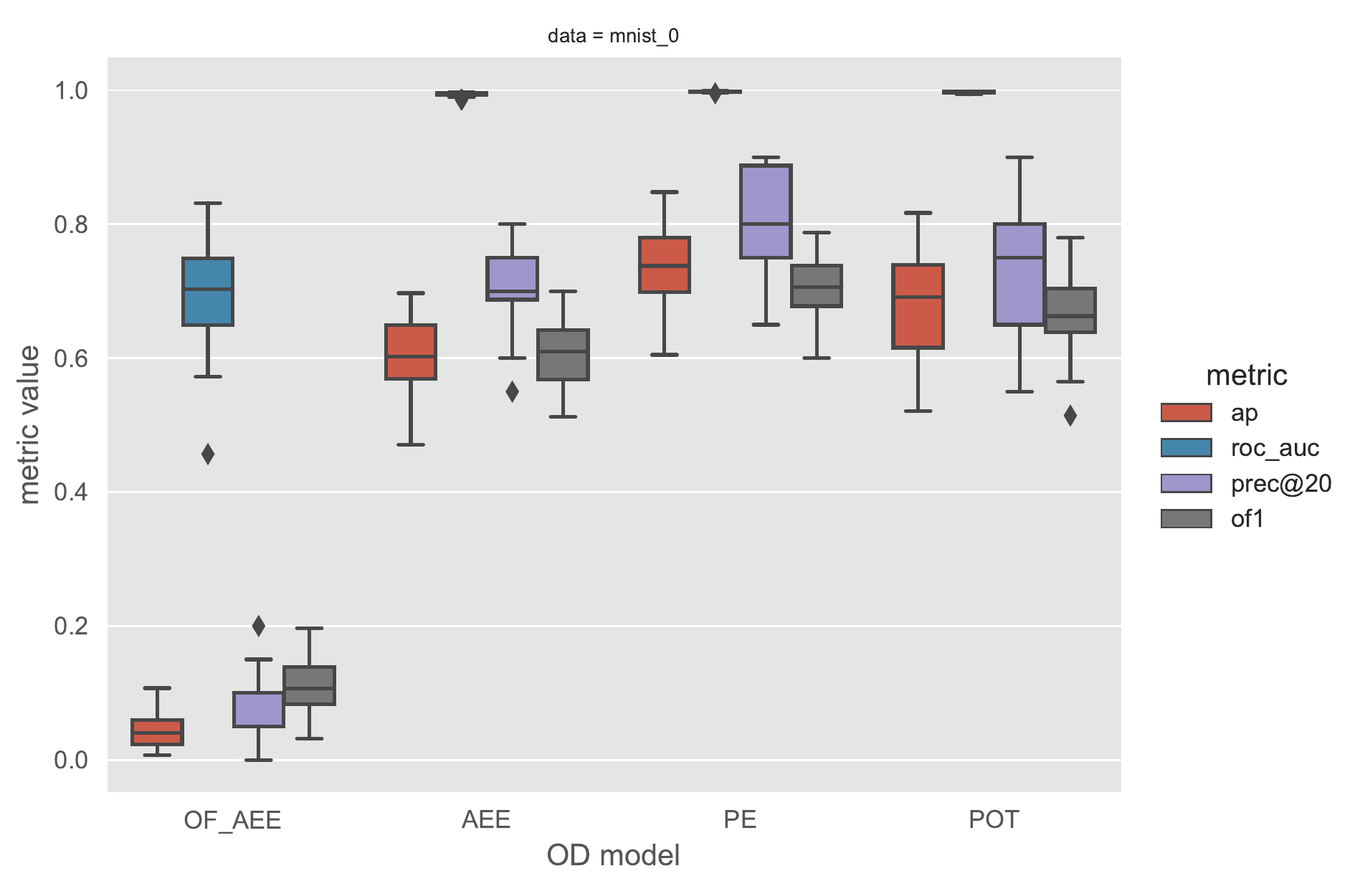}
    \caption{Evaluation of three ensemble versions: an ensemble of five
    nonregularized (overfitting) autoencoders (OF\_AEE), an ensemble of five
    regularized autoencoders (AEE), an ensemble of five POTATOES models (PE),
    and for comparison, again the single POTATOES model.}
    \label{ensBox}
\end{figure}

We now turn to a closer investigation of the effect of the partitioning of the
training data. For this, we investigate how an overfitting ensemble of
autoencoders {\em without} partitioning would perform. I.e.\ we keep all the
parameters of POTATOES, the autoencoder architecture, zero regularization, the
ensemble size of five, and the aggregation of the outlier scores of the five
ensemble member models by the max function. The only difference is that each
member of the ensemble is fitted on the full dataset and not just on an element
of a partition; this model is abbreviated by OF\_AEE. Furthermore, we include
the investigation of an ensemble of {\em regularized} autoencoders {\em
without} partitioning, abbreviated to AEE. Again, we use the same
basic autoencoder architecture as above. The only difference between OF\_AEE
and AEE is the amount of regularization. Finally, we also consider an ensemble
of five copies of the POTATOES model (i.e.\ an ``ensemble of ensembles''),
referred to as PE.

In this comparison, 30 datasets from mnist\_0 have been used. The results are
shown in Figure
\ref{ensBox}.
\begin{table}[htbp]
    \caption{The p-values for model comparisons AEE-POT and POT-PE}
    \begin{center}
        \begin{tabular}{|l| r r|}
            \hline
              & AEE-POT  & POT-PE   \\
            \hline                                       
            p & 6.855e-6 & 5.044e-4 \\
            \hline
        \end{tabular}
        \label{ensPvalues}
    \end{center}
\end{table}
First, we clearly see that the overfitting autoencoders ensemble is not
performing well, underlining that partitioning is crucial. As far as the other
models are concerned, the plot suggests a ranking of the POTATOES ensemble
coming first, followed by POTATOES, and the ensemble of regularized
autoencoders being third. For the investigations of significance the metric AP
has again been singled out. As before, neither Shapiro-Wilk nor
Kolmogorov-Smirnov object to the assumption that the paired differences between
the AP values are normal, so the paired two-sample t-test can be applied.
Table \ref{ensPvalues} displays the belonging p-values: AEE-POT is for the
comparison between the models AEE and POTATOES, and POT-PE is the comparison
between POTATOES and PE.  The values show that the ranking is significant.

\section{Conclusion}
In this paper we have introduced POTATOES, a new method for autoencoder-based
unsupervised outlier detection: For any given autoencoder architecture, it
avoids the time consuming search for the optimal neural network regularization
while still providing competitive UOD performance. In experiments we have shown
that it outperforms the regularized version of this original autoencoder
architecture, even if this regularization has been tuned using the knowledge of
the true outlier labels, which is usually not available in practice. POTATOES
doesn't use those labels because it doesn't have to tune regularization. The
only conditions for its successful application are  
\begin{enumerate*}[label=(\roman*)]
    \item that the original autoencoder is capable of overfitting the dataset
        when its regularization is removed and
    \item that the dataset is large enough so that each member $P_i$ of the
        partition contains sufficient inliers such that each autoencoder can
        still approximate any region occupied by inliers.
\end{enumerate*}

Furthermore, being an ensemble of independent autoencoders, POTATOES can be
parallelized.

Finally, we also noted that the basic underlying idea is a rather general one
of improving submanifold distance based UOD methods which need some tuning of
their ``manifold stiffness''. As such, it is not restricted to autoencoders and
is expected to work with other submanifold learning methods, too.

\end{document}